%% file: main_v1.tex
\documentclass[lettersize,journal]{IEEEtran}
\usepackage{amsmath,amsfonts,amsthm}
\usepackage{algorithmic}
\usepackage{algorithm}
\usepackage{array}
\usepackage[caption=false,font=normalsize,labelfont=sf,textfont=sf]{subfig}
\usepackage{textcomp}
\usepackage{stfloats}
\usepackage{url}
\usepackage{verbatim}
\usepackage{graphicx}
\usepackage{cite}
\usepackage{booktabs}
\usepackage{multirow}
\usepackage{makecell}
\usepackage{color}
\usepackage{CJKutf8}
\usepackage{pifont}
\usepackage{colortbl}
\newcolumntype{H}{>{\setbox0=\hbox\bgroup}c<{\egroup}@{}}
\usepackage{bm}
\usepackage{bbm}
\usepackage{pythonhighlight}
\usepackage{mathrsfs}
\usepackage{mathrsfs}
\usepackage{MnSymbol}
\usepackage{dashrule}
\usepackage{tikz}
\usepackage{mathtools}
\usepackage{amsthm}

\usepackage{adjustbox}
\usepackage{pifont}

\usepackage[utf8]{inputenc} 
\usepackage[T1]{fontenc}    
\usepackage{hyperref}       
\usepackage{url}            
\usepackage{booktabs}       
\usepackage[table]{xcolor} 
\usepackage{bm} 
\usepackage{amsfonts}       
\usepackage{nicefrac}       
\usepackage{microtype}      
\usepackage{xcolor}         
\usepackage{siunitx} 
\usepackage{subcaption}
\usepackage{wrapfig}
\usepackage{multirow}
\usepackage{multicol}
\usepackage{booktabs}       
\usepackage{graphicx}
\usepackage{amsmath}
\usepackage{array}
\usepackage[table,xcdraw]{xcolor}
\usepackage{color}
\usepackage{algorithm}
\usepackage{algorithmic}
\usepackage{tabularx}
\usepackage{hyperref}     
\usepackage{booktabs}   
\usepackage{soul}          
\usepackage{xcolor}        

\begin{document}

\title{VMU-Diff: A Coarse-to-fine Multi-source Data Fusion Framework for Precipitation Nowcasting}

\author{Chunlei~Shi, Hao~Li, Yufeng~Zhu, Boyu~Liu, Yongchao~Feng, Zengliang Zang, Hongbin Wang, Yanlan Yang, and Dan~Niu$^{\dagger}$

\thanks{This work was supported by the  Heavy Rainfall Research Foundation of China (No. BYKJ2025M14), China Meteorological Administration Xiong\textrm{'}an Atmospheric Boundary Layer Key Laboratory (No. 2025LABL-B12), and by the National Natural Science Foundation of China (62374031, 62331009), and by NSFC-Jiangsu Province (BK20240173). (\textit{$^{\dagger}$Corresponding author: Dan Niu.})}
\thanks{Chunlei Shi, Yufeng Zhu, Boyu Liu, Yanlan Yang and Dan Niu are with the Department of Automation, Southeast University, Nanjing 210096, China (e-mail: 230238514@seu.edu.cn, 17312234632@163.com).}
\thanks{Hao Li and Chunlei Shi are also with Nanjing XinDa Institute of Meteorological Science and Technology, Nanjing 210096, China.}
\thanks{Yongchao Feng is with the State Key Laboratory of Virtual Reality Technology and Systems, Beihang University, Beijing 100191, China.}
\thanks{Zengliang Zang is with the College of Meteorology and Oceanography, National University of Defense Technology, Changsha 410073, China.}
\thanks{Hongbin Wang is with the Key Laboratory of Transportation Meteorology, China Meteorological Administration, Nanjing Joint Institute for Atmospheric Sciences, Nanjing 210041, China.}
}

\markboth{Journal of \LaTeX\ Class Files,~Vol.~14, No.~8, August~2021}%
{Shell \MakeLowercase{\textit{et al.}}: A Sample Article Using IEEEtran.cls for IEEE Journals}


\maketitle

\input{sec1/0_abstract}

 
\input{sec1/1_introduction}
\input{sec1/3_method}
\input{sec1/4_experiment}
\input{sec1/5_conclusion}

\bibliographystyle{IEEEtran}
\bibliography{egbib}

\end{document}

%% file: sec1/0_abstract.tex
\begin{abstract}
Precipitation nowcasting is a vital spatio-temporal prediction task for meteorological applications but faces challenges due to the chaotic property of precipitation systems. Existing methods predominantly rely on single-source radar data to build either deterministic or probabilistic models for extrapolation. However, the single deterministic model suffers from blurring due to MSE convergence. The single probabilistic model, typically represented by diffusion models, can generate fine details but suffers from spurious artifacts that compromise accuracy and computational inefficiency. To address these challenges, this paper proposes a novel coarse-to-fine \textbf{V}ision \textbf{M}amba \textbf{U}net and residual \textbf{D}iffusion (VMU-Diff) based precipitation nowcasting framework. It realizes precipitation nowcasting through a two-stage process, \textit{i.e.}, a deterministic model-based coarse stage to predict global motion trends and a probabilistic model-based fine stage to generate fine prediction details. In the coarse prediction stage, rather than single-source radar data, both radar and multi-band satellite data are taken as input. A spatial-temporal attention block and several Vision mamba state-space blocks realize multi-source data fusion, and predict the future echo global dynamics. The fine-grained stage is realized by a spatio-temporal refine generator based on residual conditional diffusion models. It first obtains spatio-temporal residual features based on coarse prediction and ground truth, and further reconstructs the residual via conditional Mamba state-space module. Experiments on Jiangsu SWAN datasets demonstrate the improvements of our method over state-of-the-art methods, particularly in short-term forecasts.
\end{abstract}

%% file: sec1/1_introduction.tex
\section{Introduction}
\label{intro}

Precipitation nowcasting is a critical spatiotemporal prediction task in meteorology, essential for issuing timely warnings and providing guidance in sectors such as agriculture and transportation \cite{ribeiro2025flowcast}. However, forecasting precipitation over very short lead times (\textit{e.g.}, 0-2 hours) remains challenging due to the highly nonlinear and chaotic nature of atmospheric processes.

Traditional numerical weather prediction (NWP) methods, although physically grounded, are computationally prohibitive for short-term forecasting because of the complexity of simulating complete atmospheric dynamics \cite{tolstykh2005some}. In contrast, radar echo extrapolation methods \cite{han2023precipitation} provide a more efficient alternative by leveraging historical radar observations to predict future echo patterns. Nonetheless, most existing approaches rely solely on radar data \cite{wang2018predr, shi2015conv}. While radar is effective at capturing the internal structure of convective storms and current precipitation intensity, it tends to be less sensitive to early cloud dynamics and the initiation of convection. Meteorological satellite data, however, offer complementary information regarding cloud top and early convective development. Radar and satellite data exhibit distinct characteristics due to different imaging mechanisms, thereby providing complementary views. Yet, integrating these heterogeneous data sources remains challenging because of mismatched spatiotemporal resolutions and data heterogeneity.

Over the past years, deep learning has driven substantial progress in precipitation nowcasting, leading to two main categories of methods: deterministic and probabilistic models. Deterministic models (e.g., ConvLSTM \cite{shi2015conv}, TrajGRU \cite{shi2017shi}, and hybrid architectures \cite{bai2022rain, yang2022aa}) primarily employ CNNs, Vision Transformers, or their combinations to capture large-scale motion patterns. However, the use of losses such as mean square error (MSE) often results in averaging effects and blurred predictions, especially over extended lead times. In contrast, probabilistic models, typically implemented via diffusion frameworks \cite{yu2024diffcast}, are capable of generating finer details. Nevertheless, in high-resolution settings these models may produce spurious artifacts and exhibit prediction instability, which restricts their practical scalability. Moreover, while CNN-based architectures struggle to capture long-range dependencies, Transformer-based models incur quadratic complexity that further challenges high-resolution forecasting. To overcome these 
fundamental limitations, we leverage Vision Mamba, a state-space model 
that captures long-range spatiotemporal dependencies with \emph{linear 
complexity}~\cite{hu2024zigma}.

To address these challenges, we propose a novel coarse-to-fine multi-source fusion framework termed \textbf{V}ision \textbf{M}amba \textbf{U}net and residual \textbf{D}iffusion (VMU-Diff) for precipitation nowcasting. In the initial coarse prediction stage, VMU-Diff employs a deterministic approach that integrates radar and multi-band satellite data using VMSS blocks, thereby leveraging the complementary strengths of these sources to improve the capture of large-scale motion dynamics. In the subsequent fine-grained prediction stage, a conditional diffusion-based refine generator is used to model and reconstruct the residual between the coarse prediction and ground truth. This generator utilizes a Conditional Mamba State-Space (CMSS) module to adaptively refine local details and reduce prediction uncertainty. Through a structured, coarse-to-fine, multi-stage pipeline, VMU-Diff achieves a context-aware approach for accurate and scalable precipitation forecasting. In summary, our main contributions are as follows:

\begin{itemize} 
    \item  We propose a coarse-to-fine multi-source fusion framework for precipitation nowcasting, integrating radar and satellite data. It captures global motion trends through a deterministic model-based coarse prediction stage, followed by a probabilistic model-based fine-grained adjustment stage to enhance forecast detail generation.
    \item At the coarse prediction stage, we design VMU network where multi-band satellite data and radar data are integrated. And we use a Vision Mamba State-Space (VMSS) module to strengthen the forecast\textrm{'}s initial accuracy, addressing both the initiation and dissipation of precipitation.
    \item {At the fine prediction stage, we propose a conditional diffusion-based refine generator. By designing a Conditional Mamba State-Space (CMSS) block, it enables precise refinement of coarse forecast for high-resolution details and improved forecast quality with accelerated inference speed.}
\end{itemize}

%% file: sec1/3_method.tex
\begin{figure*}[ht]
\centering
\includegraphics[scale=0.52]{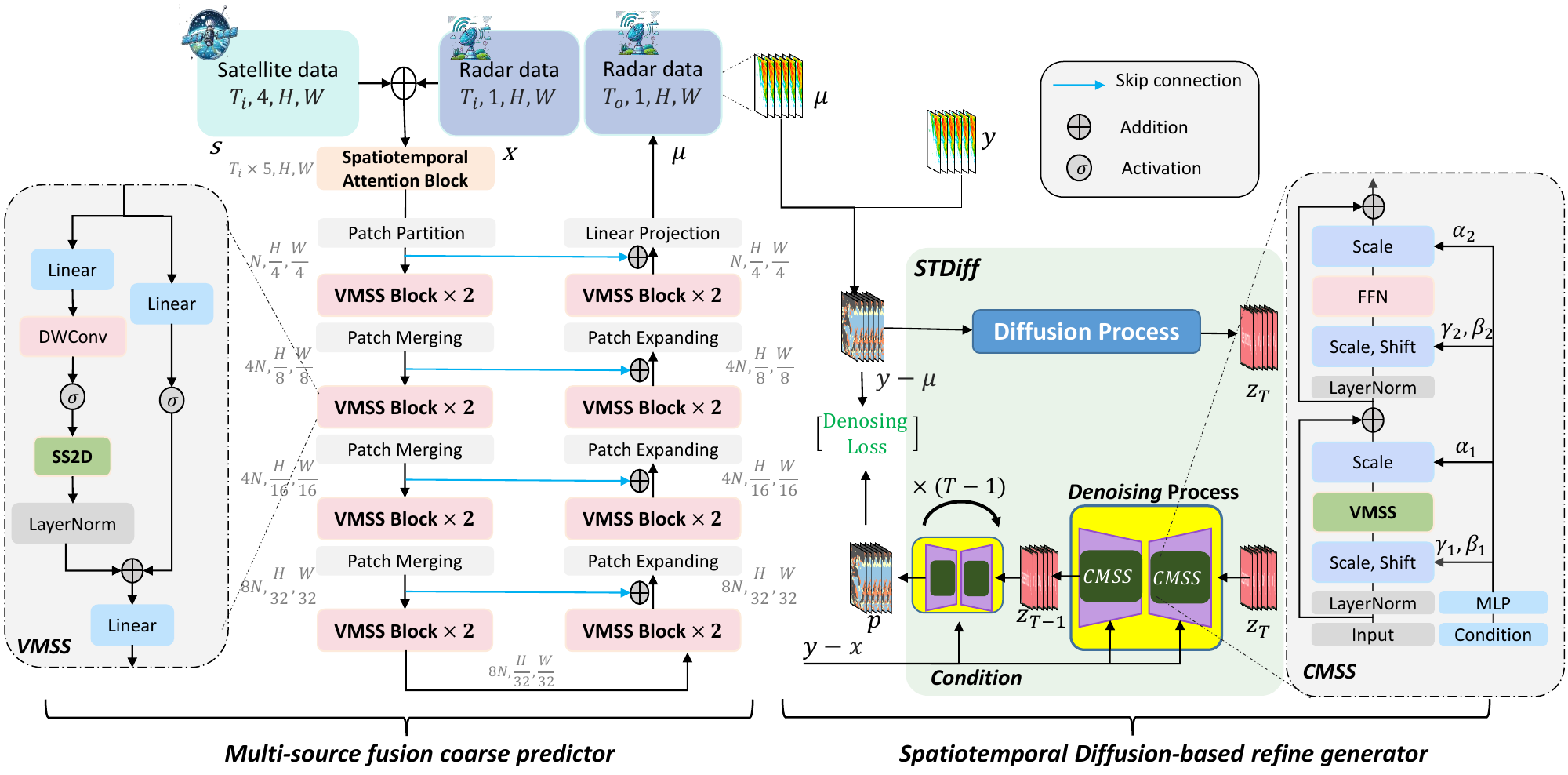}
\caption{Illustration of our coarse-to-fine multi-source VMU-Diff framework for precipitation nowcasting. The framework integrates radar and satellite data through a multi-source fusion coarse predictor and applies a spatiotemporal diffusion-based refine generator. Here, $s$ represents satellite data, and $x$ denotes radar data. Both $s$ and $x$ undergo shallow feature extraction through a spatiotemporal attention block. The extracted features are then passed through a VMU composed of VMSS Blocks, guiding the multi-source fusion process. The coarse predictions are further processed by STDiff, where a base predictor $\mathcal{P}_{\theta_1}$ generates initial coarse predictions $\mu$. During training, the difference $y - \mu$ represents the portion that the denoising STDiff needs to reconstruct, with the residual echo trend serving as guiding conditions. During inference, the framework samples the residual distribution from Gaussian noise, which is then added to the coarse prediction $\mu$ to obtain the final output $\hat{y}$.}
\label{fig2_framework}
\vspace{-5mm}
\end{figure*}
\section{Method}
\label{method}

\subsection{Overall Framework}

VMU-Diff is a coarse-to-fine framework (see Fig.~\ref{fig2_framework}) combining deterministic and 
probabilistic models for precipitation nowcasting. Given radar and satellite 
sequences $R = \{r_1, \ldots, r_n\}$ and $S = \{s_1, \ldots, s_n\}$ with 
1 km spatial and 10 min temporal resolutions, the framework predicts $m$ 
future radar frames $\hat{Y} = \{\hat{y}_1, \ldots, \hat{y}_m\}$ via:
\begin{equation}
\hat{Y} = f_{\text{VMU-Diff}}(R, S),
\end{equation}
where two stages are sequentially executed: (1) a deterministic coarse 
predictor generates $\mu$ capturing global motion, and (2) a probabilistic 
diffusion-based refine generator produces residual refinements.

\subsection{Multi-source Fusion Coarse Predictor}

\subsubsection{Spatio-temporal Attention Block (STAB)}

STAB enhances multi-source feature representation by selectively emphasizing 
spatial and temporal regions. It comprises Spatial Attention Block (SAB) 
and Channel-Temporal Attention Block (CTAB). SAB applies per-timestep 
channel pooling via $5 \times 5$ convolution to generate spatial attention 
maps. CTAB sequentially applies temporal and channel attention via fully 
connected layers. The output $X_{\text{STAB}} \in \mathbb{R}^{B,T \times C,H,W}$ 
fuses radar and satellite features for downstream processing.

\subsubsection{Vision Mamba State-Space (VMSS) Block}

To capture global weather dynamics, VMSS leverages state-space models with 
linear complexity rather than the quadratic overhead of attention mechanisms. 
The block consists of a linear embedding layer followed by dual pathways: 
one applies depth-wise convolution and SiLU activation to the SS2D module; 
the other applies SiLU directly. By omitting positional embeddings and MLP 
phases, VMSS enables dense stacking for efficient multi-source feature fusion 
while preserving long-range spatiotemporal dependencies.

The underlying SSM formulation maps input $x(t)$ to output $y(t)$ via hidden 
state $h(t)$:
\begin{equation}
h'(t) = Ah(t) + Bx(t), \quad y(t) = Ch(t) + Dx(t),
\end{equation}
where $A, B, C, D$ are learnable parameters. VMSS integrates convolutional 
operations via a Cross-Scan Module for enhanced local feature extraction 
while maintaining temporal coherency.

\subsection{Spatio-temporal Diffusion-based Refine Generator}

The refine generator takes fused features $F_{\text{fused}}$ as input and 
generates residual prediction $p = \hat{Y} - \mu$ to enhance coarse 
prediction $\mu$:
\begin{equation}
\hat{Y} = \mu + p,
\end{equation}
where $p$ is refined by the diffusion model.

\subsubsection{Residual Conditional Diffusion}

To reconstruct residuals in latent space, a conditional diffusion model 
leverages multi-source fused features as guiding conditions. The diffusion 
process is formulated as:
\begin{equation}
p(z_{0:T} \mid z_{\text{cond}}) = p(z_T) \prod_{t=1}^{T} p_{\theta}(z_{t-1} 
\mid z_t, z_{\text{cond}}),
\end{equation}
where $z_{\text{cond}}$ represents the conditioned feature from fusion 
module. Training follows standard diffusion objectives:
\begin{equation}
\mathcal{L}_{\text{refine}} = \mathbb{E}_{t,\epsilon \sim \mathcal{N}(0,I)} 
\| \epsilon - \epsilon_{\theta}(z_t, t, z_{\text{cond}}) \|^2,
\end{equation}

\subsubsection{Conditional Mamba State-Space (CMSS) Block}

CMSS incorporates both noisy input $z_T$ and condition embedding $c$ to 
adaptively refine features. The condition $c$ (via MLP) generates scaling 
$(\gamma_1, \gamma_2)$ and shifting $(\beta_1, \beta_2)$ parameters applied 
with residual skip connections:
\begin{align}
o_1 &= \gamma_1 z_T + \beta_1, \\
o'_1 &= o_1 + \text{FFN}(o_1), \\
o_2 &= \gamma_2 \text{VMSS}(o'_1) + \beta_2.
\end{align}
The final denoised output is produced via convolution on $o_2$. Parameters 
$\gamma_1, \gamma_2$ are initialized to zero to ensure stable training.

\subsubsection{Loss Function}

The total loss balances coarse and refined predictions:
\begin{equation}
\mathcal{L}_{\text{total}} = \alpha \mathcal{L}_{\text{coarse}} + 
(1-\alpha)\mathcal{L}_{\text{refine}},
\end{equation}
where $\alpha$ (set to 0.7) adjusts the contribution balance between 
deterministic and diffusion-based components.

%% file: sec1/4_experiment.tex
\section{Experiment}
\subsection{Experimental Setting}

\begin{figure*}[!t]
\centering
\includegraphics[scale=0.52]{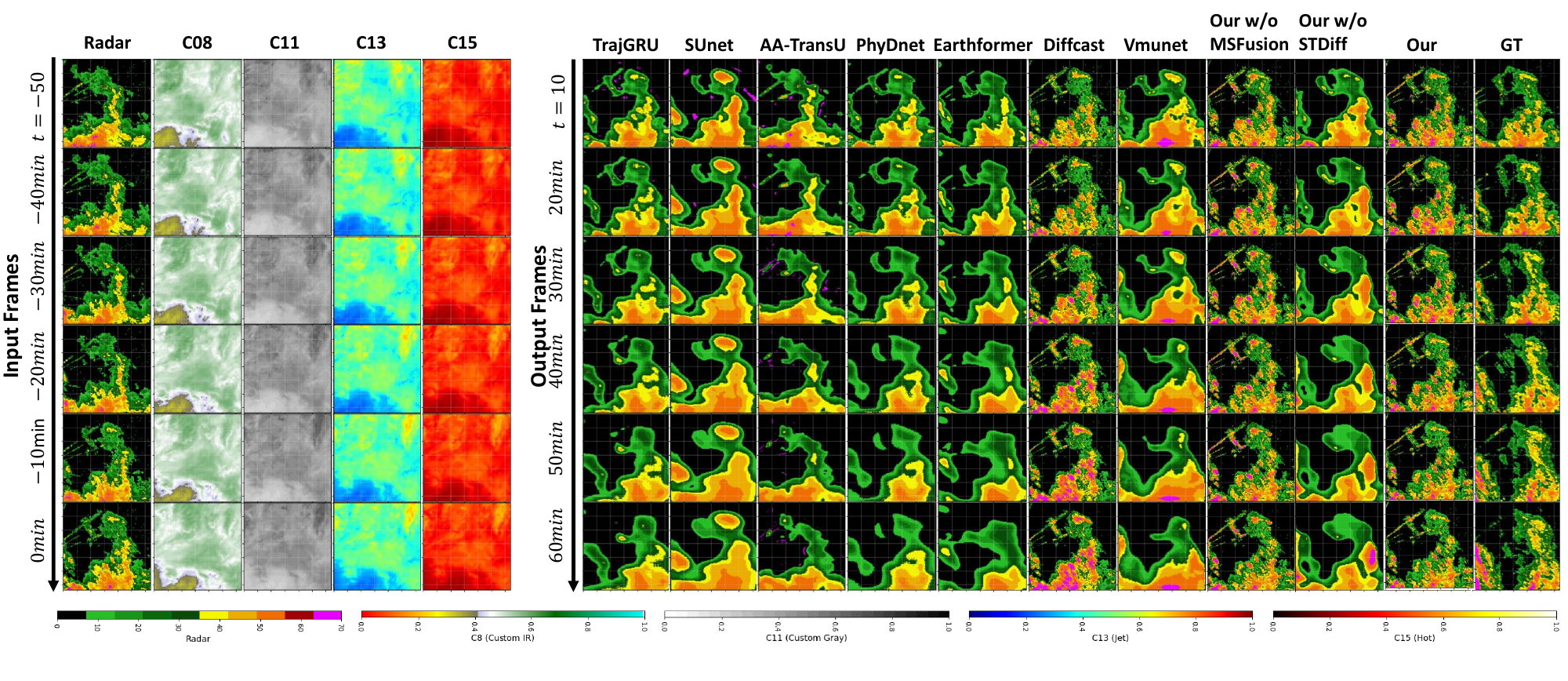}
\caption{Qualitative comparison of predicted radar echoes between VMU-Diff and other SOTA models.} \label{fig_vis}
\vspace{-2mm}
\end{figure*}

\textbf{Dataset.} The Jiangsu SWAN dataset consists of radar and Himawari-8 satellite data (infrared channels C8, C11, C13, C15) collected from June to August in 2019--2021, with a spatial resolution of 1 km and a temporal resolution of 10 minutes (300*300 pixels). Data from August 2019 to August 2021 is used for training, while data from June and July 2019 is reserved for testing.

\textbf{Evaluation Metrics.} Evaluate performance using metrics such as Critical Success Index (CSI), Heidke Skill Score (HSS), and False Alarm Ratio (FAR) across reflectivity thresholds (25, 35, 40, 45, 50 dBZ). Additionally, Structural Similarity Index Measure (SSIM) and Learned Perceptual Image Patch Similarity (LPIPS) assess the quality of generated radar imagery.

\begin{table*}[h]
\centering
\caption{Performance comparison of different deep learning models under \textbf{CSI}, \textbf{HSS}, \textbf{FAR} and Image Quality metrics. 
}
\vspace{-0.05in}
\label{tab:csi_hss_comparison}
\renewcommand{\arraystretch}{1.1}
\resizebox{\linewidth}{!}{
\begin{tabular}{c|ccccc|ccccc|ccccc|cc}
\hline
\multirow{2}{*}{Models} & \multicolumn{5}{c|}{CSI$\uparrow$} & \multicolumn{5}{c|}{HSS$\uparrow$} & \multicolumn{5}{c|}{FAR$\downarrow$} & \multicolumn{2}{c}{Image Quality} \\
\multicolumn{1}{c@{\,}|}{} & 25dBZ & 35dBZ & 40dBZ & 45dBZ & 50dBZ & 25dBZ & 35dBZ & 40dBZ & 45dBZ & 50dBZ & 25dBZ & 35dBZ & 40dBZ & 45dBZ & 50dBZ & SSIM$\uparrow$ & LPIPS$\downarrow$ \\
\hline
TrajGRU       & 0.657 & 0.597 & 0.542 & 0.450 & 0.307 & 0.700 & 0.693 & 0.662 & 0.589 & 0.436 & 0.132 &0.200 &0.237 &0.305 &0.406 & 0.331 & 0.674 \\
SmaAt-Unet    & 0.653 & 0.568           & 0.518 & 0.442 &0.347 & 0.676 & 0.650           & 0.627 & 0.572 & 0.487 & 0.228 & 0.354 & 0.411 & 0.490 & 0.590 & 0.412 & 0.683 \\
AA-TransUnet  & 0.623 & 0.568           & 0.526 & 0.455 & 0.337 & 0.661 & 0.659           & 0.641 & \underline{0.591} & \underline{0.480} & 0.166 & 0.271 & 0.312 & 0.379 & 0.478 & 0.353 & 0.506 \\
Earthformer   & 0.655 & 0.552           & 0.507 & 0.443 & 0.342 & 0.676 & 0.631           & 0.614 & 0.571 & 0.481 & 0.250 & 0.395 & 0.444 & 0.515 & 0.620 & 0.311 & 0.682 \\
DiffCast      & 0.605 & 0.476           & 0.400 & 0.315 & 0.206 & 0.516 & 0.496           & 0.464 & 0.408 & 0.306 & 0.222 & 0.286 & 0.339 & 0.397 & 0.501 & 0.265 & 0.323 \\
Vmunet        & 0.668 & 0.576           & 0.508 & 0.426 & 0.317 & 0.701 & 0.660           & 0.616 & 0.555 & 0.450 & 0.186 & 0.345 & 0.422 & 0.502 & 0.605 & 0.407 & 0.624 \\
\hline
$w/o$ MSFusion &0.666 & 0.589 & 0.534 & 0.458 & 0.348 & 0.707 & 0.675 & 0.644 & 0.589 & 0.487 & 0.180 &0.309 & 0.374 & 0.455 & 0.568 &0.277 & 0.310 \\
$w/o$ STDiff  & 0.662 & 0.565           & 0.506 & 0.428 & 0.326 & 0.693 & 0.648           & 0.613 & 0.556 & 0.461 & 0.195 & 0.352 & 0.420 & 0.499 & 0.605 & 0.430 & 0.687 \\
Our           & 0.656 & 0.575           & 0.515 & 0.443 & 0.328 & 0.690 & 0.663           & 0.628 & 0.576 & 0.466 & 0.181 & 0.295 & 0.353 & 0.417 & 0.522 & 0.284 & 0.301 \\
\hline
\end{tabular}}
\label{table:table1}
\vspace{-4mm}
\end{table*}

\textbf{Training Details.} We train the VMU-Diff framework for 200,000 iterations using the Adam optimizer with a learning rate of 0.0001. The diffusion model follows standard configurations with 1000 diffusion steps and 250 denoising steps for inference, utilizing the DDIM sampler. To balance the contributions of deterministic and denoising losses, we set the hyper-parameter \( \alpha = 0.7 \). All experiments are conducted on a Tesla V100 GPU. 

\subsection{Comparison Results}

\textbf{Visualization Comparison}. Fig.~\ref{fig_vis} presents a qualitative comparison of predicted radar echoes between VMU-Diff and several state-of-the-art (SOTA) models, including TrajGRU, SmaAt-Unet, AA-TransUNet, Phydnet, Earthformer, Diffcast, and VMUnet. From the figure, it is evident that deterministic models such as TrajGRU, SmaAt-Unet, AA-TransUNet, Phydnet, Earthformer, and VMUnet excel at capturing the overall motion trends of precipitation events. However, these models tend to produce increasingly blurry predictions as the lead time grows, and the areas of strong echo regions expand unrealistically over time. On the other hand, the probabilistic model Diffcast achieves a higher level of detail, but its predictions for strong echo locations are less accurate, often generating spurious information in these areas.

\textbf{Threshold-based Comparison}. Quantitative comparisons, shown in Tab.~\ref{tab:csi_hss_comparison}, our VMU-Diff method demonstrates superior performance on image quality metrics, achieving higher SSIM and lower LPIPS scores compared to other models. In terms of CSI, HSS, and FAR metrics, our method achieves results comparable to state-of-the-art (SOTA) methods across various thresholds. These outcomes indicate that VMU-Diff not only provides high-quality radar imagery but also delivers meaningful insights for meteorologists. 

\subsection{Ablation Study}

To evaluate the contributions of VMU-Diff\textrm{'}s components, we conducted ablation studies by removing the spatiotemporal attention block (\textit{w/o MSFusion}) and the Diffusion-Guided Refinement module (\textit{w/o STDiff}). As shown in Tab.~\ref{tab:csi_hss_comparison}, removing \textit{MSFusion} weakens the integration of radar and satellite data, leading to a decline in CSI and HSS metrics. The absence of \textit{STDiff} has a more significant impact on image quality, with a notable drop in SSIM and LPIPS scores, demonstrating its role in refining visual fidelity and enhancing the coherence of generated radar images.

Fig.~\ref{fig_vis} (rightmost columns) presents qualitative ablation results. Compared to \textit{w/o MSFusion} and \textit{w/o STDiff}, VMU-Diff produces predictions with finer spatial details and superior spatiotemporal coherence. These results underscore the complementary roles of multi-source fusion and diffusion-based refinement in the proposed coarse-to-fine framework. By integrating radar and satellite observations with conditional diffusion refinement, VMU-Diff delivers accurate and visually consistent precipitation nowcasts, providing a practical solution for operational meteorological forecasting.

%% file: sec1/5_conclusion.tex
\section*{Conclusion}
\label{sec:conclusion}

This paper presents VMU-Diff, a coarse-to-fine multi-source fusion framework for precipitation nowcasting. The framework employs a two-stage architecture: a deterministic coarse predictor that integrates radar and multi-band satellite data via Vision Mamba State-Space (VMSS) blocks, and a conditional diffusion-based refinement module that progressively enhances coarse predictions using a Conditional Mamba State-Space (CMSS) design. The VMSS block enables efficient modeling of large-scale spatiotemporal dynamics through selective state-space mechanisms, while the CMSS-guided diffusion process improves fine-grained details and reduces prediction uncertainty. 